\documentclass[conference]{IEEEtran}
\usepackage {graphicx} 
\usepackage{multirow}

\usepackage{subfig}

\begin{document}
\title{DeeperBind: Enhancing Prediction of Sequence Specificities of DNA Binding Proteins}

\author{\IEEEauthorblockN{Hamid Reza Hassanzadeh}
\IEEEauthorblockA{Department of Computational Science\\ and Engineering\\Georgia Institute of Technology\\Atlanta, Georgia 30332\\
Email: hassanzadeh@gatech.edu}
\and
\IEEEauthorblockN{May D. Wang}
\IEEEauthorblockA{Department of Biomedical Engineering\\Georgia Institute of Technology\\and Emory University\\Atlanta, Georgia 30332\\
	Email: maywang@bme.gatech.edu}

}

\newcommand{\MYfooter}{\smash{
		\hfil\parbox[t][\height][t]{\textwidth}{\centering
			\thepage}\hfil\hbox{}}}

\makeatletter
\def\ps@IEEEtitlepagestyle{%
	\def\@oddhead{\parbox[t][\height][t]{\textwidth}{
			Published in 2016 IEEE International Conference on Bioinformatics and Biomedicine (BIBM)
		}\hfil\hbox{}}%
	}

\maketitle

\begin{abstract}
Transcription factors (TFs) are macromolecules that bind to \textit{cis}-regulatory specific sub-regions of DNA promoters and initiate transcription. Finding the exact location of these binding sites (aka motifs) is important in a variety of domains such as drug design and development. To address this need, several \textit{in vivo} and \textit{in vitro} techniques have been developed so far that try to characterize and predict the binding specificity of a protein to different DNA loci. The major problem with these techniques is that they are not accurate enough in prediction of the binding affinity and characterization of the corresponding motifs. As a result, downstream analysis is required to uncover the locations where proteins of interest bind. Here, we propose DeeperBind, a long short term recurrent convolutional network for prediction of protein binding specificities with respect to DNA probes. DeeperBind can model the positional dynamics of probe sequences and hence reckons with the contributions made by individual sub-regions in DNA sequences, in an effective way. Moreover, it can be trained and tested on datasets containing varying-length sequences. We apply our pipeline to the datasets derived from protein binding microarrays (PBMs), an in-vitro high-throughput technology for quantification of protein-DNA binding preferences, and present promising results. To the best of our knowledge, this is the most accurate pipeline that can predict binding specificities of DNA sequences from the data produced by high-throughput technologies through utilization of the power of deep learning for feature generation and positional dynamics modeling.
\end{abstract}


\IEEEpeerreviewmaketitle

\section{Introduction}
DNA binding proteins are key components of different cell processes including transcription, translation, repair, and replication machinery. Protein-DNA interactions play important roles in all these components and the processes thereof. One of the important protein-DNA interactions that are vital for expression of genes in the cells is the interaction between transcription factors (TF) and their corresponding binding sites. TFs are widely present in the cells to the extent that their coding genes have been estimated to comprise 5-10\% of the genes in Eukaryotes \cite{Ho06}. Finding binding sites associated with a transcription factor can help predict genes that it regulates, annotate its functions as well as functions of genes it regulates, understand regulatory processes in biological systems, identify causal disease variants and, more importantly, design pharmaceutical drugs that promote or prevent expression of target genes. This calls for accurate molecular techniques to pinpoint the locations on DNA where these factors bind, which in itself requires the ability to measure precisely the binding affinity between these molecules. With recent advances in high-throughput technologies in the past decade several in-vivo \cite{Chipseq,Chippet} and in-vitro \cite{Ber09,HTSELEX,HiTSFLIP} techniques have been invented and upgraded to address this important and yet challenging task. Unfortunately, none of these methods are able to generate results that are interpretable by biologists, but instead each generates a large volume of noisy, erroneous and low-resolution measurements for tens of thousands sequence probes. As a result, the outcome of such experiments need to be processed through downstream analysis pipelines to elicit useful information. In this study, we use data from Protein Binding Microarrays (PBM) experiments to evaluate our proposed method. PBM \cite{Ber09} is a recent in-vitro high-throughput technology that can massively measure relative binding preferences of DNA probes for a given transcription factor. The binding preference in a PBM experiment is directly related to the measured spot intensities which are scanned and recorded for later analysis. We apply our method to the data produced by a set of PBM experiments and try to predict the binding preferences for the test probes.\\

Deep learning has recently gained momentum due to its success in improving the previously recorded state-of-the-art performance measures in a wide range of domains \cite{AlphaGO,ResNet,DeepBind} including motif elucidation. DeepBind \cite{DeepBind} is perhaps the most notable success story for an application of deep learning to a challenging problem in biological domain which was proposed to address the shortcomings of classical tools in a scalable and efficient way through its deep convolutional architecture. To the best of our knowledge \cite{DeepBind} has reported the most accurate prediction results among all the available models proposed to date. Here we propose DeeperBind, a novel doubly-deep model for prediction of sequence specificities of transcription factors and show that our model surpasses DeepBind on available popular metrics. Moreover, notwithstanding our focus in this article, we report previously published performance measures on the same datasets from four pioneering classical methods: RankMotif++ \cite{Rankmotif}, PERGO \cite{PERGO}, KmerHMM \cite{Kmerhmm} and, Seed \& Wobble \cite{Ber06b}. In the remainder of this article, we give a brief description of DeepBind as the baseline model and outline our proposed model, both in section \ref{method}. Then, in section \ref{results}, we benchmark our pipeline on two different metrics and finally conclude the paper in section \ref{conclusion} and provide possible future directions to extend our method.

\section{MATERIALS AND METHODS}
\label{method}
Models based on convolutional neural networks (CNNs) have become an integral part of the winner models in almost every image processing contest these days. This was perhaps the major motivation for transforming DNA sequences into image-like modalities rendering them amenable to deep models. In fact, the same way that pixel values in a 2D image depend on the neighboring pixels, the set of nucleotides in a DNA locus that has a specific function (such as a promoter) are not just a random assortment of nucleotides and hence can be treated as a 1D image. Not only that, the set of nearby pixels (nucleotides) often exhibit stationary patterns which obviate the need for deploying fully connected layers until the last stages of deep pipelines. Additionally, the convolution operation is fundamentally similar to scanning sequences with PWM (position weight matrix) filters which was once the basis for promising solutions that characterize motifs. Therefore, it is favorable to take advantage of these powerful models in a deep way to benefit from the recent optimization techniques and pattern recognition paradigms.\\

\subsection{DeepBind}
DeepBind \cite{DeepBind} was the first deep convolutional method ever designed to address the need for accurate characterization of motifs for protein targets. Despite its name, it employs only one convolution layer followed by a non-linear thresholding, a max-pooling layer and one/two fully connected layer(s) to estimate the intensity of input probes. In summary, it computes a binding score $f(s)=net_W(pool(rect_b(conv_M(s))))$ using the four stages mentioned above with an end-to-end trainable architecture. Here, the convolution stage ($conv_M$) scans a set of $4\times m$ motif detectors, $M_k$, across the sequences much like applying a PWM of length $m$ but without requiring the coefficients to be log-odd ratios. Note that in order to be applicable to varying length sequences, the pooling layer has to take the maximum over the complete sequence so that the fully connected layer receives inputs with consistent dimensions. This by itself deprives the model from benefiting the positional information that exists in the feature maps generated from the convolution layer. Finally, they use recent deep learning techniques such as the dropout regularization, use of momentum in optimization and calibrating parameters to boost the training efficiency. Overall, DeepBind owes its promise to its three unique features 1) training several convolution filters (possibly with different lengths, through zero padding) each being capable of learning some aspect of the DNA-Protein interaction, 2) efficient training algorithms, regularization techniques and non-linearities, thanks to the maturity of the field in recent years and, 3) benefiting from the complete set of experiment data due to its scalable and GPU-trainable architecture.\\

Despite its clever design, DeepBind lacks the ability to capture the positional dynamics of probe sequences by the indirect assumption that there exists at most one motif in each probe. This preconception can mislead the training process, for instance, in the not so rare situations where several moderately good motifs in a probe have produced a high binding affinity due to the individual contributions made by each. In such cases, the model tries to tune motif detectors' (kernels') weights such that all but one motif are penalized while the remaining one is overweighted due to the high affinity of the whole probe. Even for a probe with only one motif, the positional location may matter due to the technological limitations. For example, a motif which is located towards the ends of a probe attracts the proteins with a different affinity than the same motif when placed in the middle of the probe. Finally, convolutional neural networks fall short capturing the long-term (motif level) and short-term (nucleotide level) dependencies within probes. The long-term dependency is particularly important due to the existence of possibly multiple attractive probe sub-regions where as the short-term dependency is important because of the interactions between the higher-order structure of proteins and the DNA motifs. 

\subsection{Recurrent Neural Networks}
\begin{figure}[]
	\centering
	\subfloat[]{\includegraphics[width=2.7in]{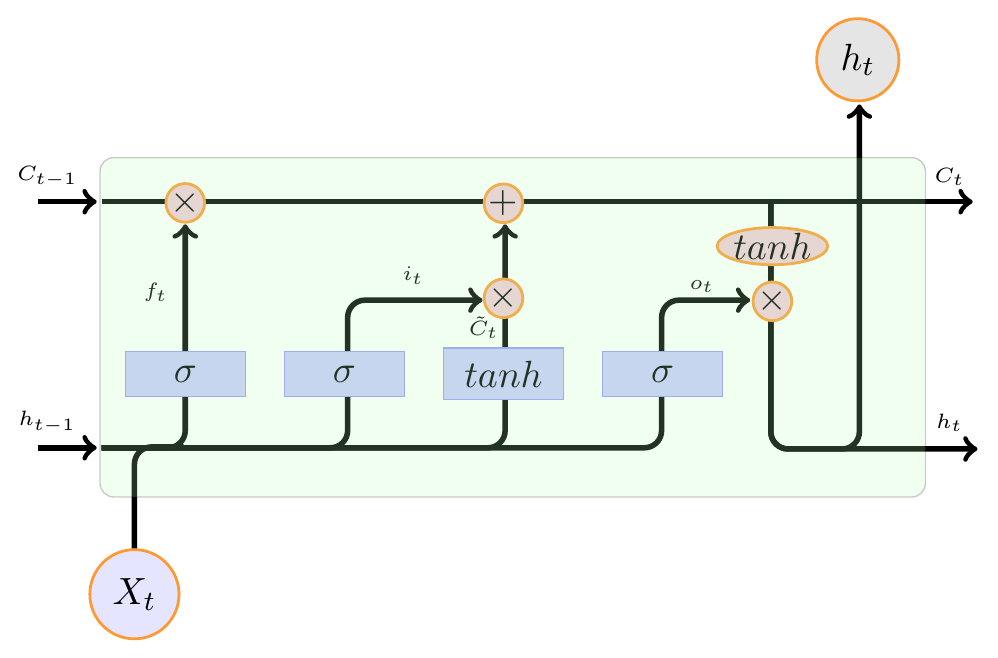}%
		\label{lstm}}
	\\
	\subfloat[]{\includegraphics[width=3.2in]{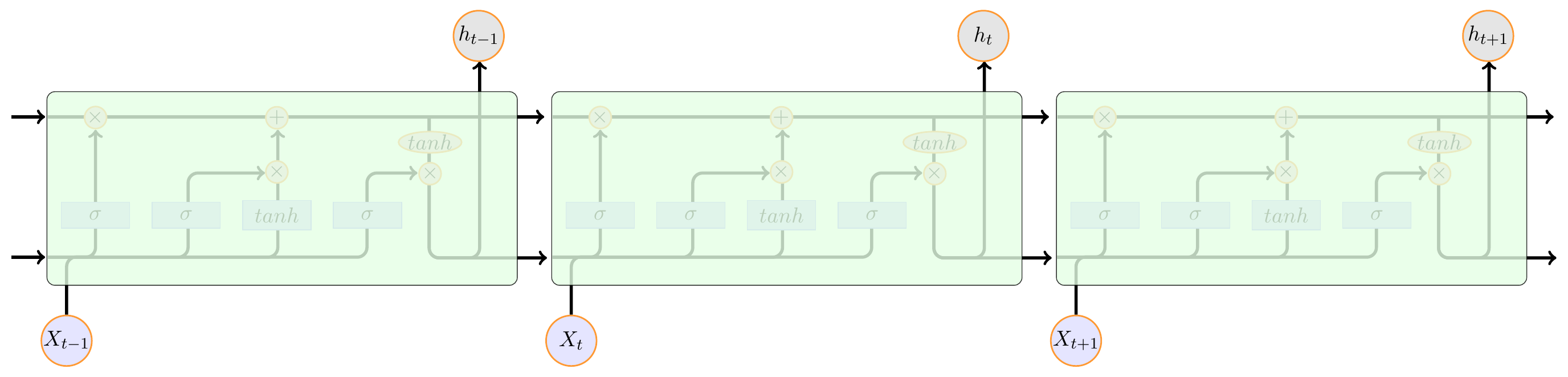}%
		\label{lstmseq}}
	\caption{Recurrent neural networks. (a) Diagram of an LSTM memory cell used in this paper. Small pink circles denote pointwise operations, arrows indicate vector transfer, merged arrows mean concatenation of vectors and split arrows indicate vector copying operation. (b) concatenation of a few LSTM cells to build a deep LSTM network.}
	\label{recurrent}
\end{figure}

Recurrent Neural Network (RNN) models are well known for their deepness in time/position. These models were introduced even before the spatial convolutional models
\ and have been widely explored in different domains for perceptual applications. The principle behind RNNs is that sequential signals which exhibit stationary features in time can be learnt via a recurrent network that unrolls as we proceed through. The unrolled units share the same set of parameters as they learn the same task but at different time points. Despite its rich literature, RNNs failed to surpass shallow models due to their significant limitation in propagating back the error gradients towards distant time points, an issue which is often referred to as the "vanishing gradient" effect. That is, the gradients tend to become prohibitively small as they reach distant past as a result of which the long-range temporal dependencies become increasingly hard to capture. In 1997, Hochreiter et. al. \cite{LSTM} proposed a new class of RNNs, the long short term memory networks (LSTMs), that were able to efficiently solve the vanishing gradient problem by introducing a gating mechanism in a well-behaved and differentiable way. The gating strategy makes each LSTM block behaves like a memory unit and thereby allows the network to decide when and what to remember/forget as the input is presented to the model at each time step. This is achieved through four single neural network layers (as opposed to the single layer in regular RNNs) that are connected to each other in a special way (Figure \ref{lstm}). LSTM networks have become recently popular after being utilized in large-scale tasks with complex temporal state dependencies such as speech recognition and language translation tasks (See \cite{Sut14} 
\ for an example.)\\

Several variation of LSTM networks have recently been proposed which differ from each other mostly by their gates and internal connections. Figure \ref{recurrent} depicts one popular LSTM architecture which is used in our proposed pipeline. 
The gates can be considered as knobs through which the cell state ($C_{t-1}$) is selectively manipulated from the beginning to the end of the unit as the signal passes through. The gating mechanism is performed through the sigmoid layers that act on the current input ($x_t$) and the previous hidden state ($h_{t-1}$) with their parameters respectively tied across the unrolled units. The output of each gating layer is element-wise multiplied by the corresponding signals to forget, emphasize or add relevant information to the cell-state that is subsequently passed to the next unit, all in a differentiable way. Finally, aside from the deepness in time, it is sometimes favorable, as is the case in this work, to stack extra LSTM layers atop. In that case, the second LSTM layer will take as input the previous layer's hidden state.\\

\subsection{DeeperBind}
Here we introduce DeeperBind, a novel doubly-deep model (and hence the name), that has the ability to deal with deficiencies of the model described earlier. We seek to add a positional dimension to the core design of the DeepBind by including recurrence into its model. As CNN and LSTM have been shown to be complementary in their modeling capabilities \cite{LRCN2,LRCN}, it is desired to combine these two to achieve synergistic improvements in prediction of protein-DNA binding specificities. Our objective in this work is to leverage from CNN as a visual feature extractor, and LSTM as a recurrent model that recognizes and synthesizes temporal dynamics of sequential tasks. A recurrent architecture serves two purposes. First, it facilitates learning temporal/positional dynamics of the target task. Specifically, by recruiting an LSTM network we are interested to capture both long term (motif level) and short term (nucleotide level) dependencies. Note on the latter that convolution- and PWM-based approaches assume independence between adjacent nucleotide positions which does not hold in general. Nucleotide-level dependency is manifested in a variety of forms such as motifs with indels (insertion/deletion of nucleotides). As we will illustrate shortly, by using a recurrent structure we can directly and indirectly account for the former and the latter type of positional dynamics, respectively. Second, integration of RNN with CNN makes handling variable-length inputs possible, and therefore data that are produced by different platforms can be readily used in a unified framework to achieve a more accurate model. Our design is end-to-end trainable and can exploit GPU for speed gains.\\

\begin{figure}[!htbp]
	\centering
	\includegraphics[width=8.5cm ,trim=4cm 3cm 1cm 3cm,clip]{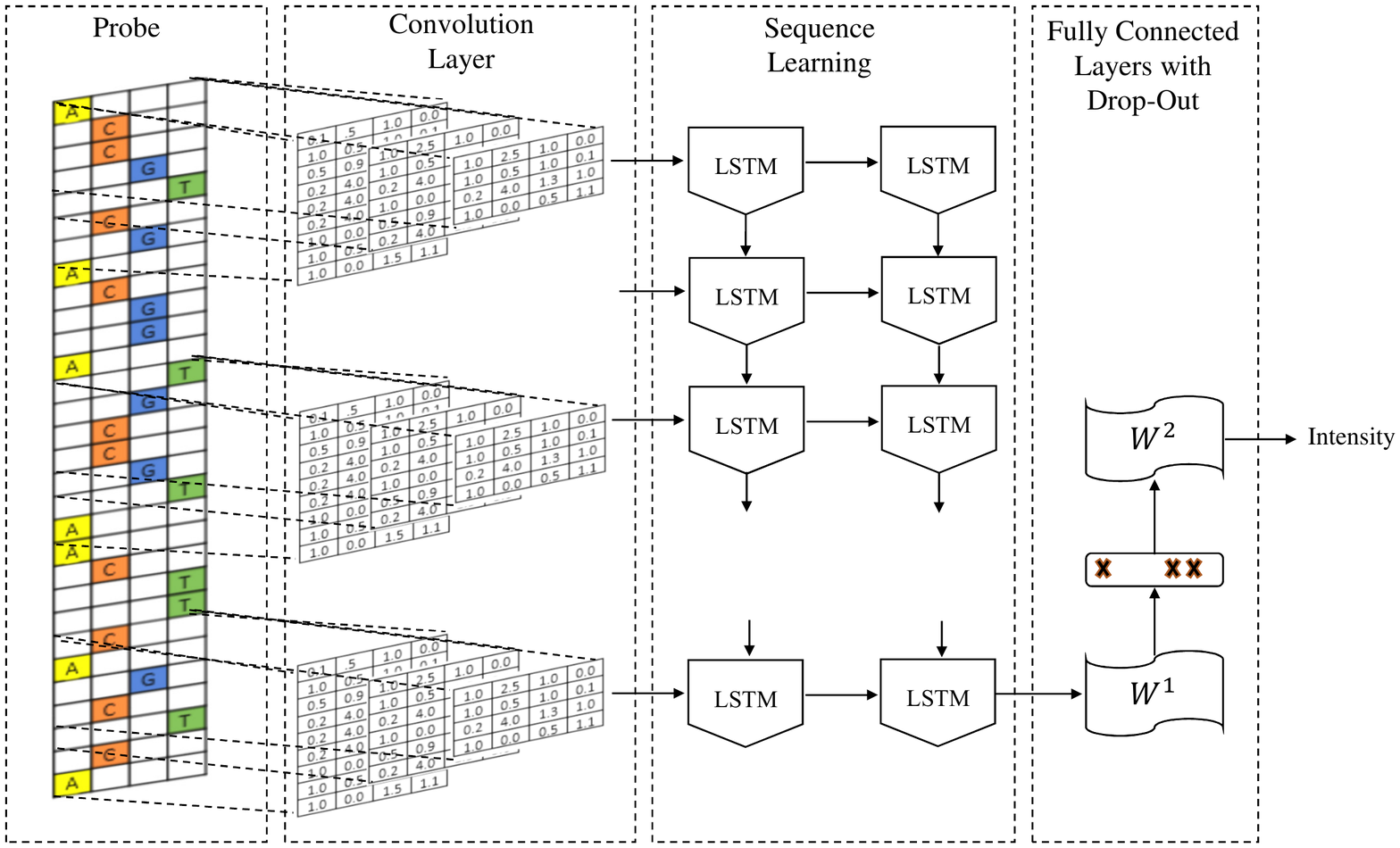}
	\caption{Block diagram of DeeperBind. The input sequences are first represented as 2D binary matrices via one-hot coding. A convolutional layer generates the feature map by applying several PWM-like filters followed by rectified linear units. No pooling layer is used. Two stacks of LSTM layers then capture the sequential dependencies of the sub-motifs on probes. }
	\label{diagram}
\end{figure}

Figure \ref{diagram} depicts the overall block diagram of the proposed pipeline. In the first step, each probe sequence is converted into a $4\times L$ one-hot coded binary matrix ($L$ is the probe length) and the intensity values are normalized. Then, we feed in the pre-processed probes into a convolutional layer followed by rectified linear units, to map them into intermediate feature vectors through parameterized non-linear transformations. Note that we omit the pooling layer which is often used in today's convnet architectures, to avoid losing positional information. Note also that we set the convolution stride step to one knowing that the higher level modules (LSTM layers) can efficiently deal with the increased redundancy. This is necessary as we want to model the positional dynamics of the probes in a nucleotide-level granularity. In the next step, we use one/two layer(s) of LSTM where each LSTM block in the first layer will receive the local features extracted from the locus of attention on the DNA and encodes its own interpretation regarding the overall contributions of the past history into its hidden state. This interpretation, in turn, is passed on to the next LSTM blocks located above and to the right of it and so on. Once the last nucleotide observed, the last unrolled LSTM block makes the final decision on the goodness of the probe, based on the processed feedbacks coming from the immediate neighbors which is an integration of all the history in an appealing way. Finally, the outcome of the LSTM network is presented to a network of fully connected layers with at most one hidden layer and drop-out regularization (see Table \ref{grid} for details) to predict the binding preference of each probe.\\

\section {Results}
\label{results}
\begin{figure}[]
	\centering
	\subfloat[CEH-22]{\includegraphics[height=2in,trim=0 1cm 0 1cm,clip]{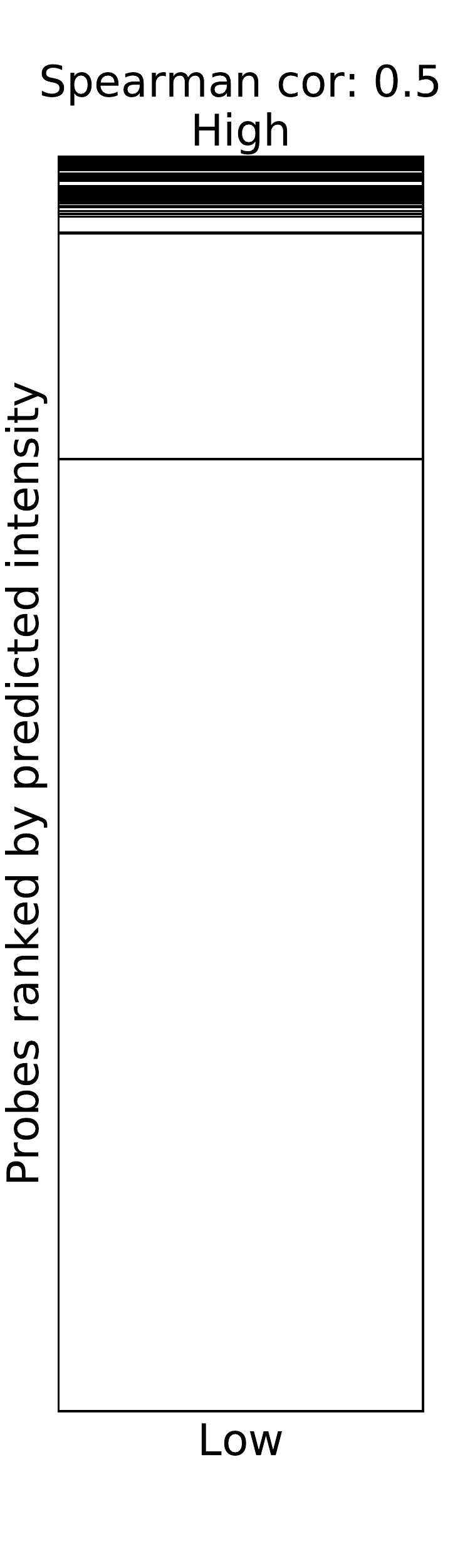}\includegraphics[height=2in,trim=0 1cm 0 1cm,clip]{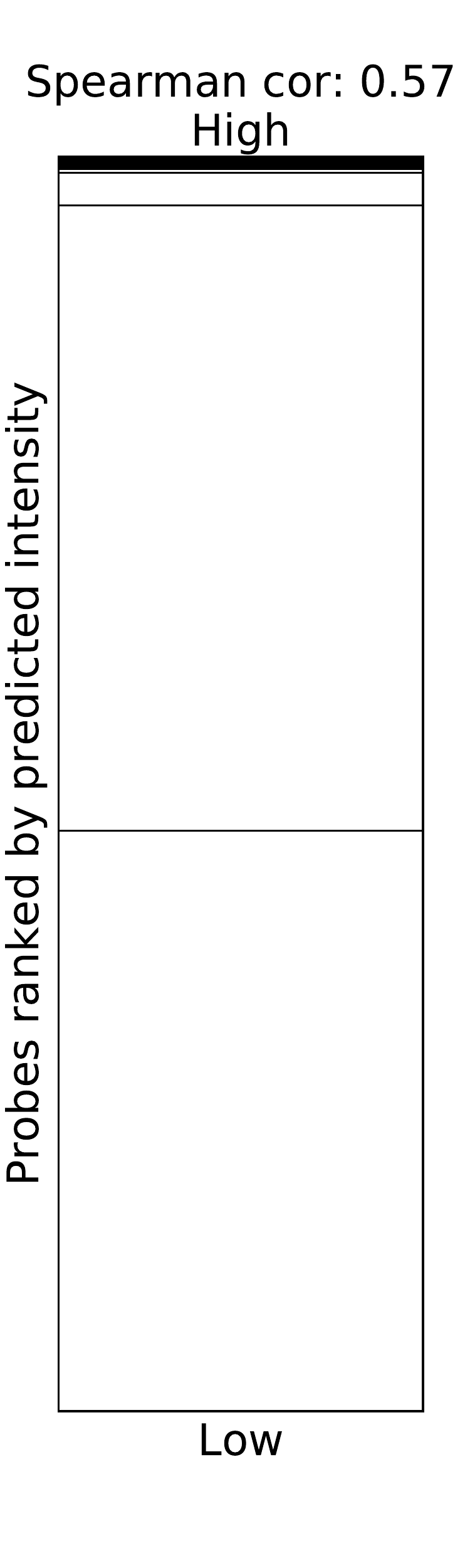}%
		\label{1d_CEH-22}}%
	\hfill
	\subfloat[Oct-1]{\includegraphics[height=2in,trim=0 1cm 0 1cm,clip]{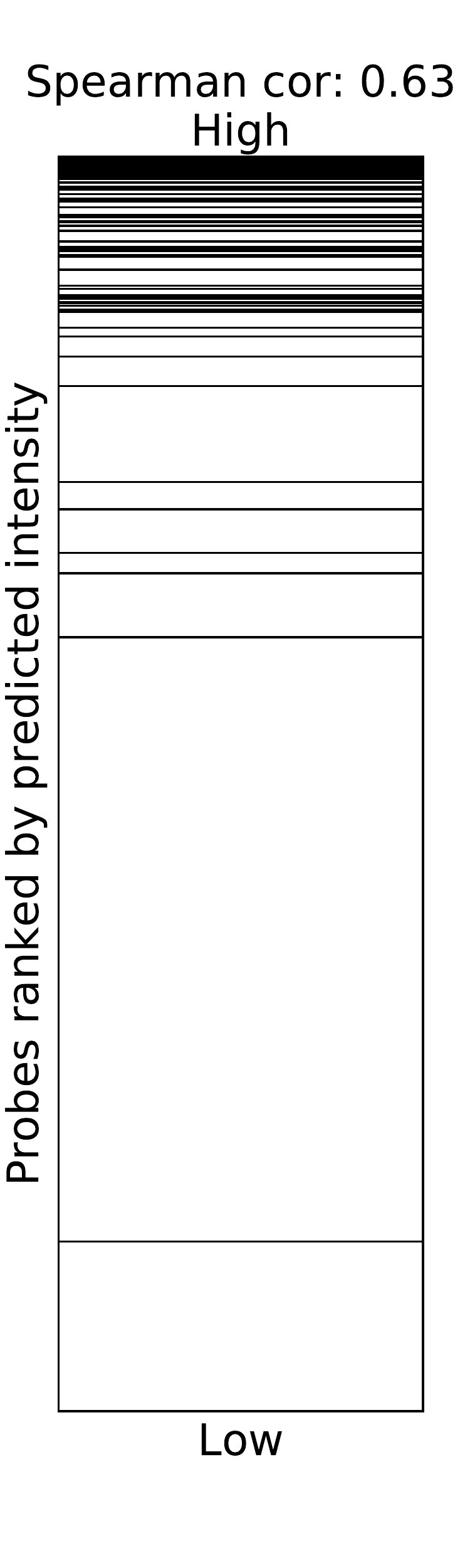}\includegraphics[height=2in,trim=0 1cm 0 1cm,clip]{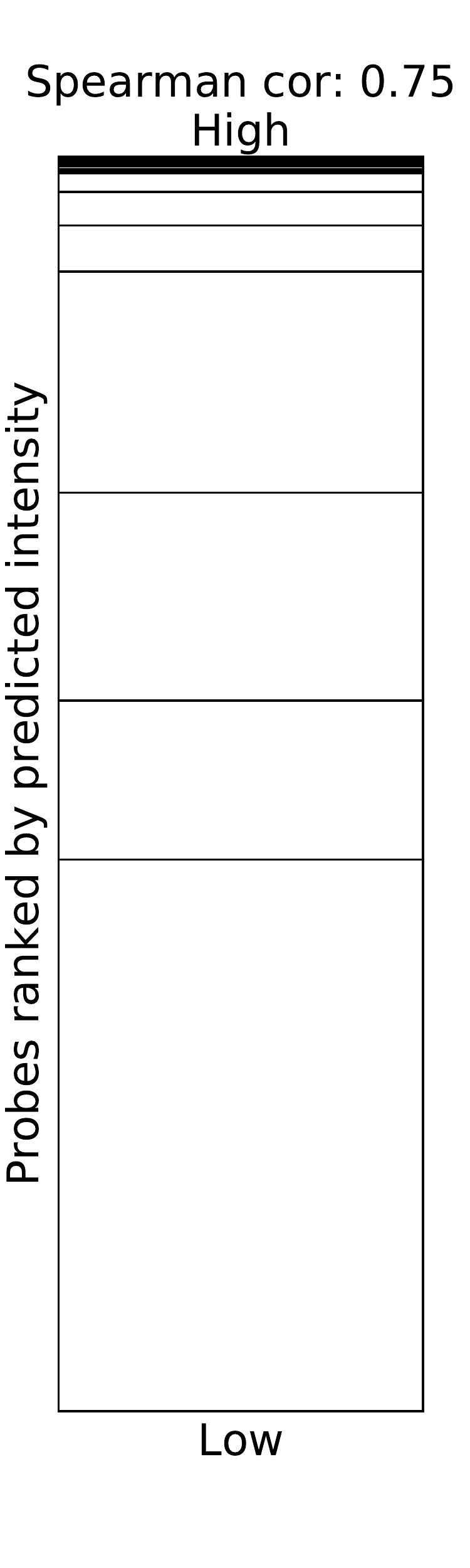}%
		\label{1d_Oct-1}}\\
	
	\caption{The predicted rank of the top 100 positive probes (black lines) in array \#1 for the DeepBind (left) and the DeeperBind (right) per each TF: a) CEH-22 and b) Oct-1.
	}
	\label{1d_comparison_regression}
\end{figure}

To make a fair comparison between our proposed work and the baseline in \cite{DeepBind}, we [re-]implemented and trained both models on the same training sets and used similar training procedures. We initialized some of the hyper-parameters (such as the architecture of fully connected layers, type of non-linear thresholding, number of motif kernels and etc.) empirically, based on the current practices. For parameters that are sensitive to small changes (i.e., the learning rate, learning rate decay, weight decay, drop-out, size of mini-batches and the LSTM network architecture) we looked for the best configuration using a grid-search strategy as illustrated in Table \ref{grid}. The models were trained on PBM experiment data retrieved from the UniProbe database \cite{UniProbe}. Specifically, we downloaded results of two independent assays (called array \#1 and \#2 that are conducted independently based on a different set of deBruijn sequences) for two popular transcription factors, namely CEH-22 and Oct-1. These datasets are commonly used for evaluation purposes in the literature \cite{Kmerhmm,Rankmotif}. Each assay contains between 40,000 and 42,000 probes with their measured intensities. Note that, these measurements are often noisy due to the limitations of the technology. As with the other published related studies, for each TF, we used the first array (which we divided into 70\% training and 30\% validation) to train each model and the second one to measure its performance. We used five motif finding kernels of length 11 for the convnets and the RMSProp algorithm \cite{RMSProp} to optimize the overall networks.\\
\begin{table} 
	\centering
	\caption{List of parameters and their corresponding range of values used in the grid search.}
	\label{grid}
	\begin{tabular}{|c|c|c|}
		\hline
		\multirow{2}{*}{Parameter} & \multicolumn{2}{c|}{Method}                                                 \\ \cline{2-3} 
		& DeepBind & DeeperBind                                                              \\ \hline
		Learning rate              & \multicolumn{2}{c|}{1e-2, 1e-3, 1e-4}                                       \\ \hline
		Learning rate decay        & \multicolumn{2}{c|}{1e-7, 1e-4, 0.0}                                        \\ \hline
		Weight decay               & \multicolumn{2}{c|}{1e-5, 0.0}                                              \\ \hline
		LSTM architecture          & -   & \begin{tabular}[c]{@{}c@{}}30, 20, 30:20,\\ 10:10, 10:20\end{tabular} \\ \hline
		Drop out                   & \multicolumn{2}{c|}{0, 0.2, 0.5}                                            \\ \hline
		Batch size                 & \multicolumn{2}{c|}{40, 100}                                                \\ \hline
	\end{tabular}
\end{table}
For each dataset we trained both the DeepBind and the DeeperBind on the training sets by minimizing the negative log-likelihood loss of the predicted probe intensities. 
\ We selected the settings that yielded the highest Spearman coefficients on the validation sets and used those for the subsequent analyses and bench-markings. Upon completion of this stage, we noticed that for all datasets there exist at least one setting with the highest score in which the drop-out is not playing any role, which could be explained by the abundant probes that obviate the need for drop-out regularization and reduce the chance of overfitting. In other words, the enormous number of probes assayed by PBM as a high-throughput technology, allows us to design models with increasingly high model capacities without being worried too much about the over-fitting problem. In light of that, we examined the capacity of each model in learning the underlying regression task. To that end, Figure \ref{1d_comparison_regression} visualizes the predicted rank of the the top 100 positive probes in the first (i.e. trian) array for each experiment. Note that traditionally a probe is marked as positive if its intensity exceeds the threshold $m+4\sigma$ where $m$ is the median of all the intensities and $\sigma$ is the median absolute deviation of all probes' normalized intensities divided by 0.6745 \cite{Rankmotif}. As clearly seen in the figures, binding preferences predicted by our method is well concentrated towards the top of the charts which means that it can successfully learn most discriminating patterns inside the training data that are good predictors of the measured intensities.\\

\begin{figure}
	\centering
	\subfloat[CEH-22: DeepBind]{\includegraphics[height=2in  ]{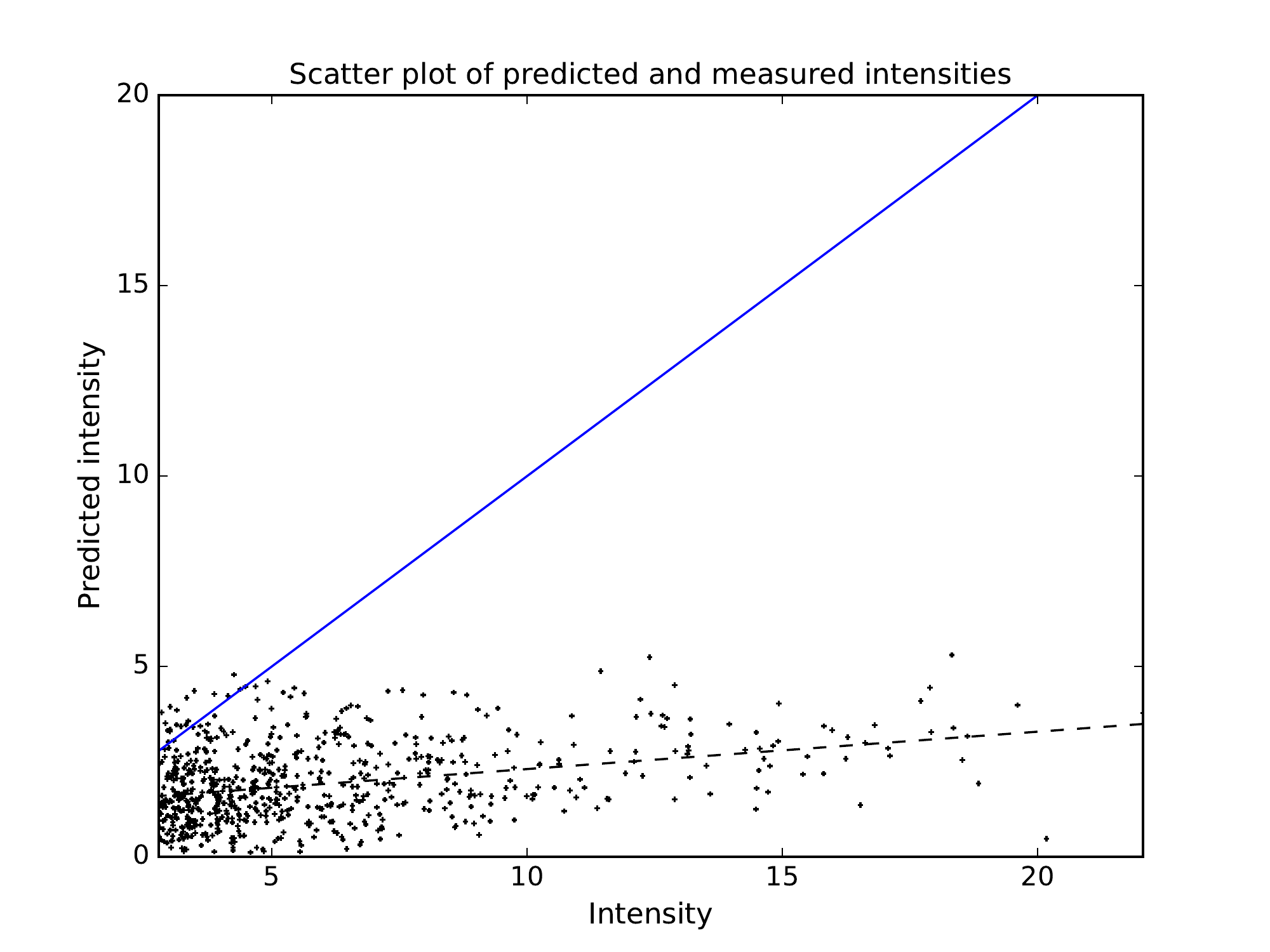}%
		\label{2d_CEH-22-deepbind}}\\
		\subfloat[CEH-22: DeeperBind]{\includegraphics[height=2in ]{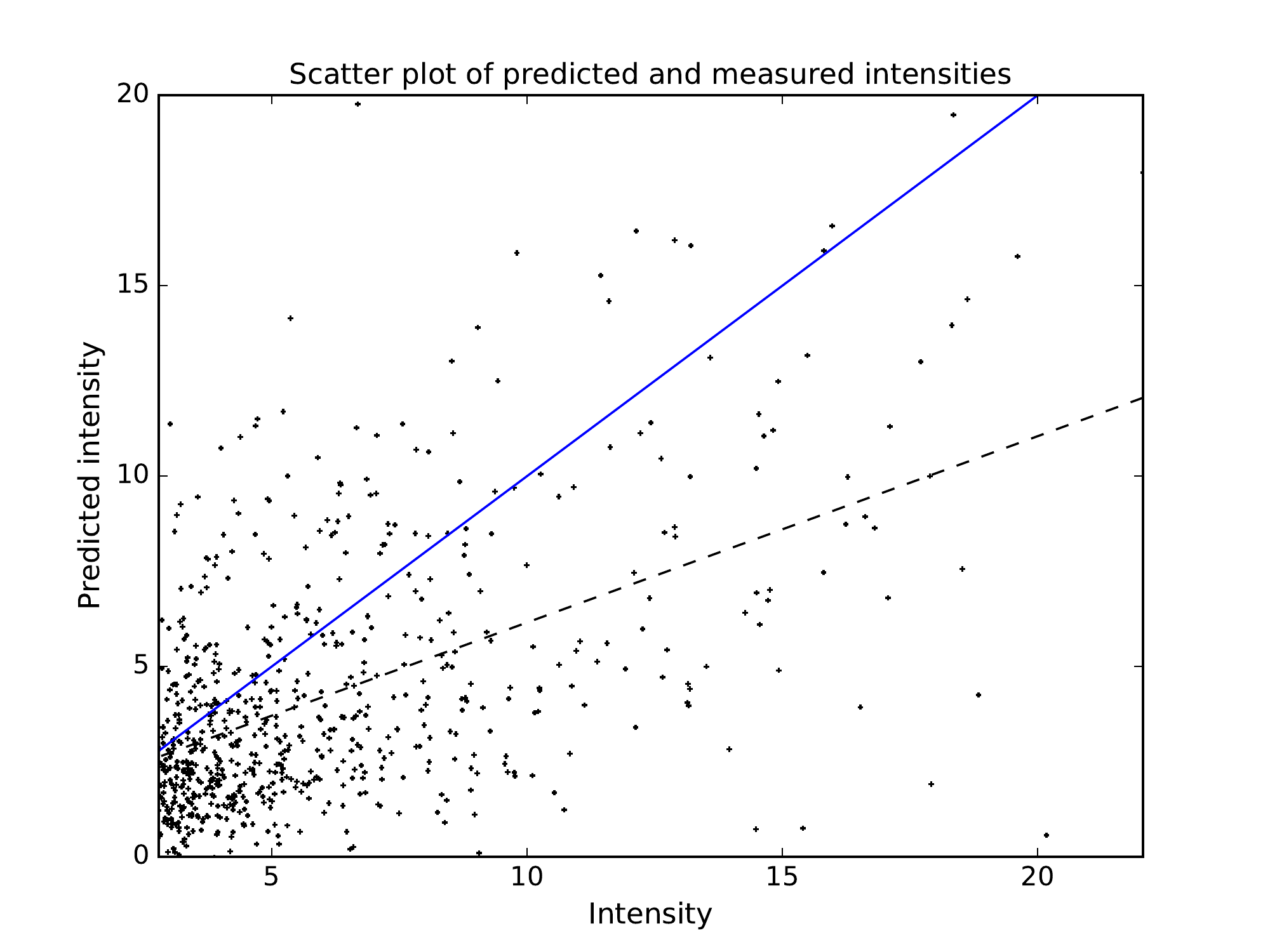}%
		\label{2d_CEH-22-deepberind}}\\
	\subfloat[Oct-1: DeepBind]{\includegraphics[height=2in ]{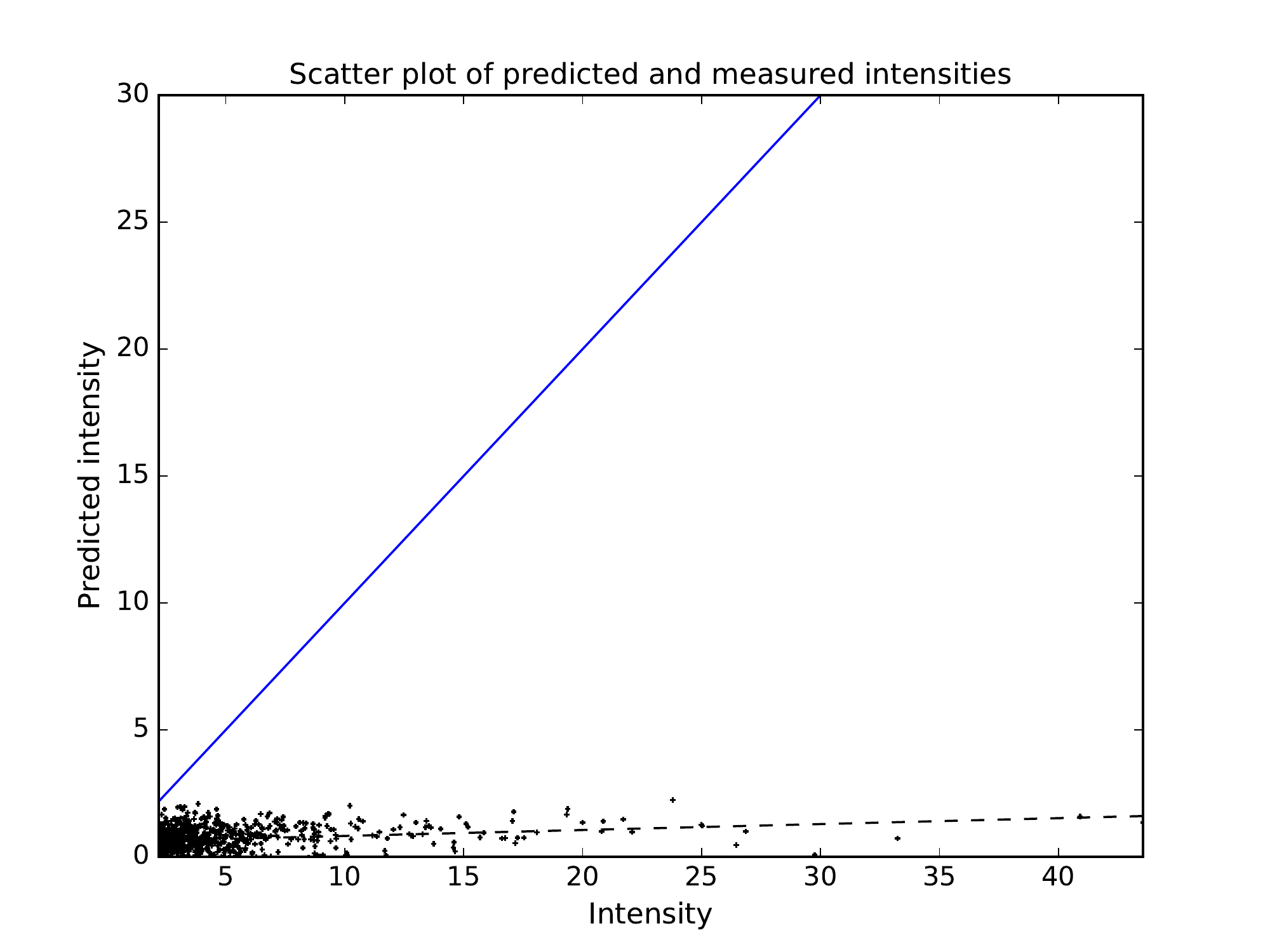}%
		\label{2d_Oct-1-deepbind}}\\
	\subfloat[Oct-1: DeeperBind]{\includegraphics[height=2in ]{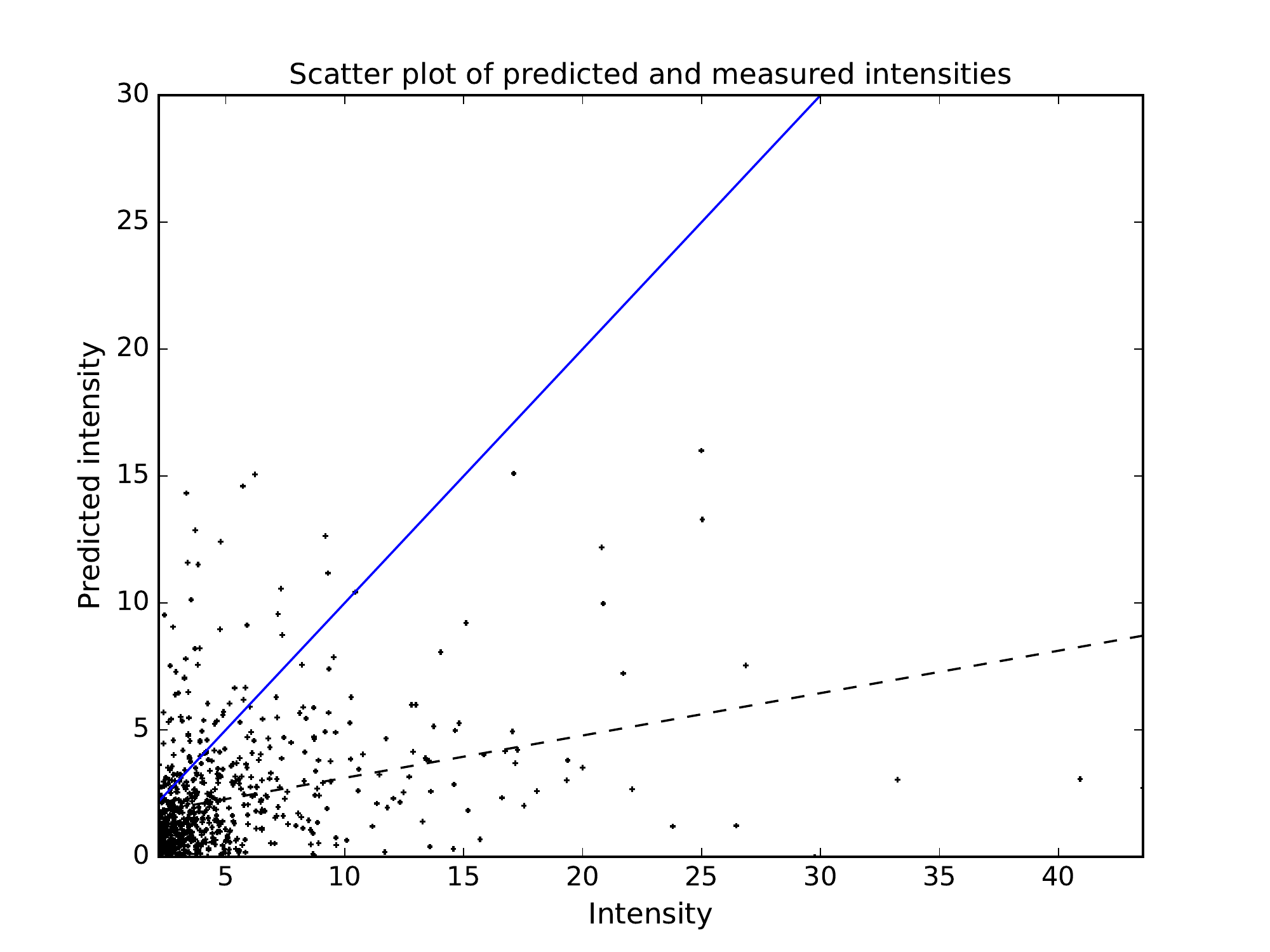}
		\label{2d_Oct-1-deeperbind}}\\
	
	\caption{Scatter plot of predicted and measured intensities: DeepBind vs DeeperBind. a) CEH-22 and b) Oct-1.
}
	\label{2d_comparison_regression}
\end{figure}

To assess the generalization capability of each method we computed the Spearman rank correlation between the measured probe intensities and the predicted ones using the trained models. Here we report this measure for each dataset and method in Table \ref{spearman_results}. We also included the same statistics for the pioneering shallow models from \cite{Kmerhmm,Rankmotif}. According to this table, DeeperBind is consistently outperforming the other methods by a large margin with DeepBind being the runner-up. Moreover, Figure \ref{2d_comparison_regression} shows the scatter plot of the predicted and measure intensities 
on the test arrays, using the same models. A similar conclusion can be made from the scatter plots in the figure. More specifically, the regression lines estimated for the predictions made by our method depicts a trend that is closer to the identity line (blue line) compared to the baseline.\\

\begin{figure}[]
	\centering
	\subfloat[CEH-22]{\includegraphics[width=2.5in,trim=1cm 0cm 1cm 0cm,clip]{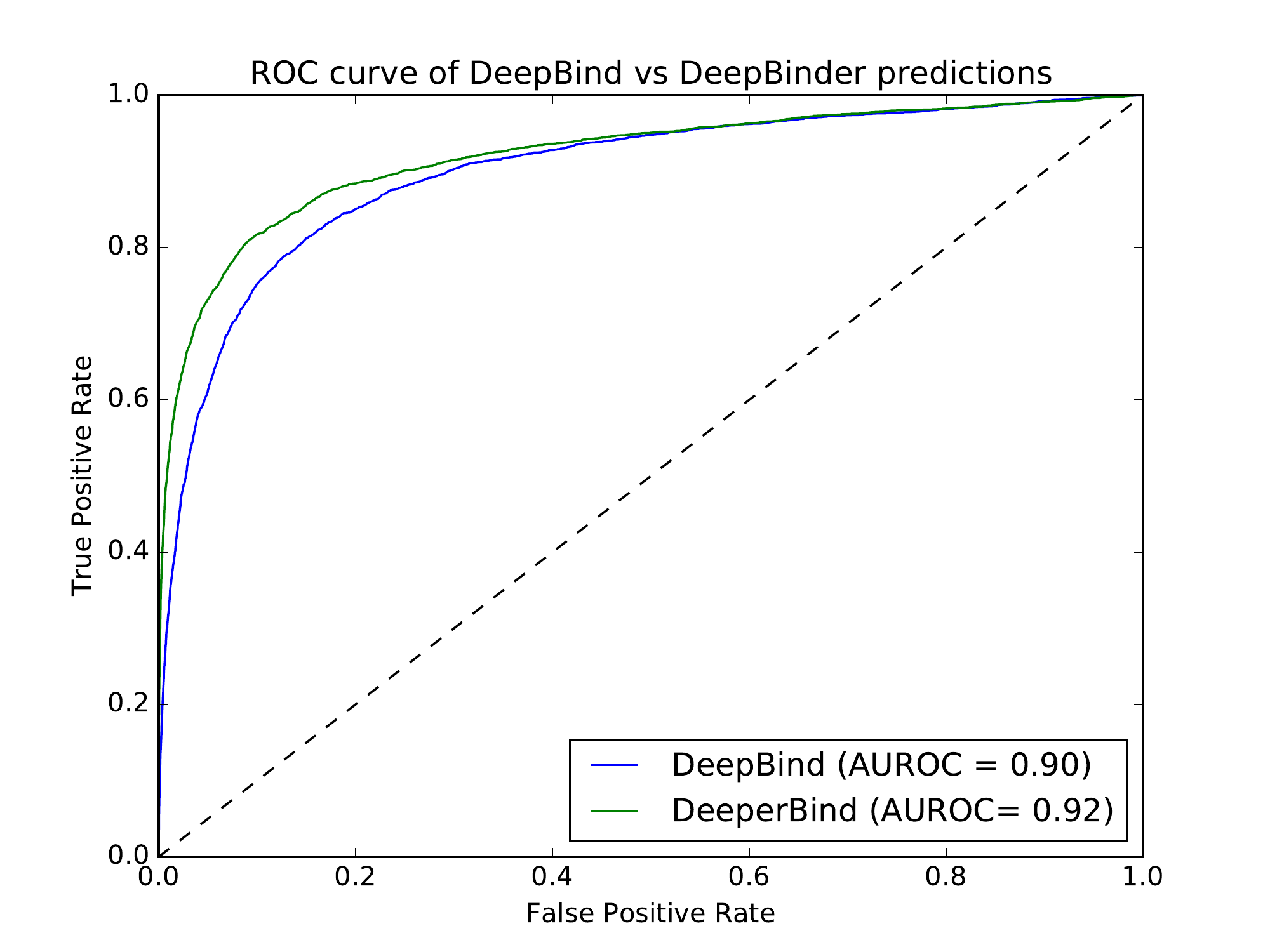}%
		\label{roc_CEH-22}}\\%
	\subfloat[Oct-1]{\includegraphics[width=2.5in,trim=1cm 0cm 1cm 0cm,clip]{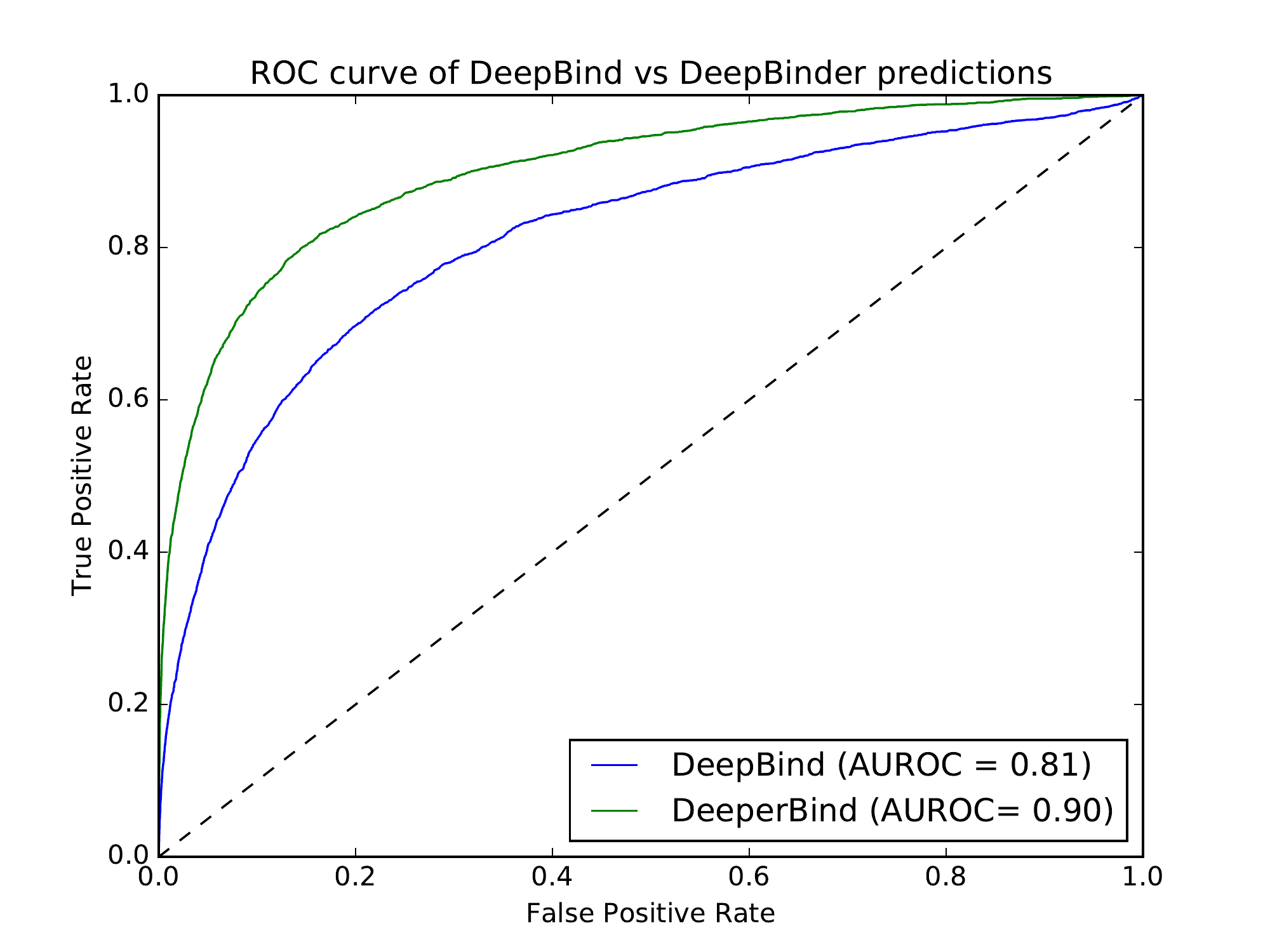}%
		\label{roc_Oct-1}}%
	\caption{Receiver Operating Characteristic (ROC) curves on array \#2. Samples are marked as positive if their measured intensity exceed $m+4\sigma$ where $m$ is the  and $\sigma$ is the median absolute deviation of all probe normalized intensities divided by 0.6745.
		}
	\label{roc}
\end{figure}

\begin{table}
	\centering
	\caption{Spearman rank correlation coefficients on array \#2 for the most accurate shallow learners PERGO, RankMotif++, Seed \& Wobble and KmerHMM (according to \cite{Rankmotif,Kmerhmm}) as well as the DeepBind \cite{DeepBind} and the DeeperBind. 
	}
	\label{spearman_results}
	\begin{tabular}{|c|c|c|c|c|c|c|}
		\hline
		\multirow{2}{*}{\textbf{TF}} & \multicolumn{6}{c|}{\textbf{Method}}        \\ \cline{2-7} 
		& PRG & RKM & S\&W & KHM & DBD & DEBD\\ \hline
		CEH-22                            & 0.28   &\bf{0.43}  & 0.28  & 0.31 & 0.40 & \bf{0.43}   \\ \hline
		Oct-1                      &  0.27   & 0.29  & 0.21  & 0.36 & 0.49 & \bfseries{0.60}   \\ \hline
	\end{tabular}
\end{table}
Finally, we conducted an experiment to examine the classification performance of the trained models. We used the same optimized models and applied them to the test arrays again but this time, instead of using probes' measured intensities we put the samples into two classes, namely the positive (for positive probes) and the negative (for the rest of probes) classes. Figure \ref{roc} depicts the ROC curves along with the area under curve of the ROC for the predictions made on the test arrays. According to this figure, DeeperBind shows a better generalization behavior despite its higher model complexity. Furthermore, for each array, we recorded the true positive rate at the 1\% false positive rate (99\% specificity). Table \ref{1_percent_tpr} demonstrates the true positive rate of the predictions for the same datasets and settings. Based on the table, DeeperBind is surpassing the other models for again for both datasets. 
\begin{table}[]
	\centering
	\caption{True positive rates at 1\% false positive rate on array \#2. The numbers in parentheses designate the number of positive probes in each array.  Columns corresponding to the shallow methods: PERGO, RankMotif++, Seed \& Wobble and KmerHMM are copied from \cite{Kmerhmm,Rankmotif}.}
	\label{1_percent_tpr}

\begin{tabular}{|c|c|c|c|c|c|c|}
	\hline
	\multirow{2}{*}{\textbf{TF}} & \multicolumn{6}{c|}{\textbf{Method}}        \\ \cline{2-7} 
	& PRG & RKM & S\&W & KHM & DBD & DEBD\\ \hline
	CEH-22 (3490)              & 0.2   &0.33  & 0.25  & 0.32 & 0.32 & \bf{0.52}   \\ \hline
	Oct-1  (3107)              &  0.27   & 0.24  & 0.20  & 0.31 & 0.18 & \bfseries{0.39}   \\ \hline
\end{tabular}
\end{table}

\section{Conclusion}
\label{conclusion}
In this article we proposed a new approach for predicting DNA binding affinity of proteins to the DNA probes using the most successful deep learning techniques, the long short term memory (LSTM) and the convolutional neural networks (CNN). Through extensive assessments and comparisons it was shown that the new design beats the most accurate prediction tools that we are aware of, often by a large margin. The promise of this work is mostly due to 1) LSTM's ability in capturing positional dynamics of the intermediate level features that are simply lost in the baseline method due to the pooling layer and 2) in its efficiency in handling a large number of redundant features generated by the deep convnets. Contrary to DeepBind, the only current deep pipeline for prediction of binding preferences, our model is capable of dealing with varying-length sequences by exploiting LSTM layers and there is no need for any pooling layer as it removes the positional dimension of the intermediate features. As a natural extension to the proposed work, we are interested to explore the capability of DeeperBind in uncovering motifs on DNA sequences by building deep motif models and to benchmark its performance comprehensively, particularly in the presence of heterogeneous data produced by different in-vivo and in-vitro high-throughput technologies.\\

\bibliographystyle{plain}
\bibliography{Paper}

\end{document}